\documentclass[times,twocolumn]{article}
\pdfoutput=1 
\usepackage{framed,multirow}
\usepackage[utf8]{inputenc}
\usepackage{amsfonts}
\usepackage{bm}
\usepackage{graphicx}
\usepackage{booktabs}
\usepackage{hyperref}
\usepackage{caption}
\usepackage{algorithm}
\usepackage{algorithmic}
\usepackage{etoolbox} 
\usepackage{authblk}

\begin{document}
\title{Class-Splitting Generative Adversarial Networks}
\author[1]{Guillermo L. Grinblat}
\author[1]{Lucas C. Uzal}
\author[1]{Pablo M. Granitto}
\affil[1]{CIFASIS, French Argentine International Center for Information and Systems Sciences, UNR-CONICET, Rosario S2000EZP, Argentina}
\date{(Under consideration at Pattern Recognition Letters)}
\maketitle

\begin{abstract}
Generative Adversarial Networks (GANs) produce systematically better quality samples when class label information is provided., i.e. in the conditional GAN setup. This is still observed for the recently proposed Wasserstein GAN formulation which stabilized adversarial training and allows considering high capacity network architectures such as ResNet. In this work we show how to boost conditional GAN by augmenting available class labels. The new classes come from clustering in the representation space learned by the same GAN model. The proposed strategy is also feasible when no class information is available, i.e. in the unsupervised setup. Our generated samples reach state-of-the-art Inception scores for CIFAR-10 and STL-10 datasets in both supervised and unsupervised setup.
\end{abstract}

\section{Introduction}
The irruption of Generative Adversarial Nets (GAN) \cite{Goodfellow2014} produced a great leap for image data generation. Samples were generated simply by applying a neural network transform to an input random vector sampled from a uniform distribution. There is no need for any Markov chains or unrolled approximate inference networks during either training or generation of samples \cite{Goodfellow2014}. GAN based generative models did not take long to reach impressive image quality \cite{Radford2015, Salimans2016, Zhao2016, Mao2016} at least for some specific datasets. 

However, current GAN models cannot produce convincing samples when trained on datasets of images with high variability, even for relatively low resolution images. On the other hand, it is observed that sample quality improves when class information is taken into account in a conditional GAN setup \cite{Mirza2014,Odena2017}. These findings suggest that it is hard to learn a multimodal distribution from a smooth transform of a uniform (or Gaussian) distribution and that providing categorical class information to the generator alleviates this problem.

Our proposal is inspired in two observations. First, as mentioned above, conditioning generation with categorical class labels with high level of abstraction improves image quality. Second, as early observed in \cite{Goodfellow2014,Radford2015}, the adversarial network pair learns an useful hierarchy of representations in an unsupervised setup. We propose to exploit the same representation space learned by the GAN model in order to generate new class labels with a high level of abstraction. This is done by applying a simple clustering method in this representation space. By conditioning generation with this new class labels the model is able to generate better samples. This can be done either when prior class information is available or not.


The main contributions of the present paper are\footnote{The source code is available at \url{https://github.com/CIFASIS/splitting_gan}}:
\begin{itemize}
 \item We propose a method for increasing the number of class labels during conditional GAN training based on clustering in the representation space learned by the same GAN model. We base our implementation on the more stable Wasserstein GAN formulation \cite{Arjovsky2017b, Gulrajani2017}.
 \item We show an effectively way of adapting networks architecture to handle this increasing number of classes.
 \item We show that this \textit{splitting} GAN strategy improves samples quality both in the supervised and unsupervised tasks, reaching state-of-the-art Inception scores for CIFAR-10 and STL-10 datasets in both tasks.
\end{itemize}
\section{Background}
In the original GAN formulation \cite{Goodfellow2014}, the generator is a neural network that transform noise input $\bm{z} \sim p(\bm{z})$ into fake samples and the discriminator $D$ is a neural network with a single scalar output with a sigmoid activation. This output is interpreted as the model probability for its input being a real image from the dataset distribution against being a fake image sampled by generator $G$. The discriminator $D$ is trained using the standard binary classification formulation by minimizing the binary cross-entropy between fake and real distributions. On the other hand, the generator $G$ is simultaneously trained to mislead the discriminator. This is accomplished in practice by updating $G$ parameters minimising the same loss but with fake samples tagged with a `true' label \cite{Goodfellow2014}. 

In other words, the discriminator is updated by gradient descent over a negative log likelihood loss
\begin{equation}\label{eq:d_loss}
L_D = -\mathop{\mathbb{E}}_{\bm{x} \sim  \mathbb{P}_r}[\log(D(\tilde{\bm{x}}))] - \mathop{\mathbb{E}}_{\bm{z} \sim p(\bm{z})}[\log(1-D(G(\bm{z})))],
\end{equation}
while the generator minimizes
\begin{equation}\label{eq:g_loss}
L_G = - \mathop{\mathbb{E}}_{\bm{z} \sim p(\bm{z})}[\log(D(G(\bm{z})))].
\end{equation}

The main issue in the original GAN formulation was the instability of the training process that made very hard to improve architectures and to scale up to bigger images.
In \cite{Radford2015} a deep convolutional architecture was proposed for both generator and discriminator which presents some degree of stability for adversarial training. This work was the first one producing convincing image samples for datasets with low
variability (Bedrooms and Faces) and relatively low resolution (64x64). However, standard GAN formulation fails to generate globally consistent samples when trained on datasets with high variability like ImageNet.
\subsection{AC-GAN}

In order to tackle datasets with high variability, Odena et al. \cite{Odena2017} proposed to improve the quality of the generated samples by adding more structure to the GAN latent space and an auxiliary classifier. This approach requires the dataset to include class labels, i.e. to work in a supervised setting. The generator receives the noise vector $z$ and also the selected label $c$ so that the generated sample is a function of both. Furthermore, the discriminator has, in addition to the usual objective, the task of correctly classifying the real and generated samples (through an auxiliary classifier). The generator is optimized not only to deceive the discriminator, but also to generate fake samples that minimize the auxiliary classifier error, i.e. to produce well class-defined samples.
\subsection{WGAN}
In order to address the problem of instability in GAN training, Arjovsky et al. in a series of works \cite{Arjovsky2017, Arjovsky2017b} proposed a reformulation of the function to be optimized. They argue that the original loss function presents discontinuities and vanishing gradients with respect to generator parameters. Instead, they proposed a distance for distributions known as Earth-Mover distance or Wasserstein-1, which captures the cost of transporting mass in order to transform one distribution into the other. From this distance they derive the WGAN loss function for the minimax objective
\begin{equation}
\mathop{\textup{min}}_{G} \mathop{\textup{max}}_{D\in\mathcal{D}} \mathop{\mathbb{E}}_{\bm{x} \sim  \mathbb{P}_r}[D(\bm{x})] - \mathop{\mathbb{E}}_{_{\bm{z} \sim p(\bm{z})}}[D(G(\bm{z}))]
\end{equation}
where $D$ (called \textit{critic} in WGAN formulation) is not anymore a binary classifier and is restricted to be in the set $\mathcal{D}$ of 1-Lipschitz functions. Again, $z$ is a noise vector sampled from a simple distribution (uniform or Gaussian distribution). The Lipschitzness of $D$ was imposed by weight clipping in this first version of WGAN.

The importance of Arjovsky's contribution lies on a gain in the robustness of the adversarial training process and a reduction in the frequency of the mode collapse phenomenon. Furthermore, the proposed loss function correlates well with the observed sample quality as opposed to the original GAN loss which gives little information about training progress.
\subsubsection{WGAN-GP}
An improved version of WGAN was recently proposed by Gulrajani et al. \cite{Gulrajani2017}. They found that the weight clipping can cause convergence problems in some settings and propose to enforce the Lipschitz constraint on the critic $D$ by penalizing its gradient's norm.

The penalty term is computed over a set of random points $\hat{\bm{x}}$ uniformly sampled from straight lines between real and fake sample points. Naming as $\mathbb{P}_{\hat{\bm{x}}}$ the latter distribution, the new loss can be written as
\begin{eqnarray}
L &=& \mathop{\mathbb{E}}_{\tilde {\bm{x}} \sim  \mathbb{P}_g}[D(\tilde{\bm{x}})] - \mathop{\mathbb{E}}_{\bm{x} \sim  \mathbb{P}_r}[D(\bm{x})] \nonumber \\
&&+ \lambda \mathop{\mathbb{E}}_{\hat{\bm{x}} \sim  \mathbb{P}_{\hat{\bm{x}}}}\big[(\|\nabla_{\hat{\bm{x}}}D(\hat{\bm{x}}) \|_2-1)^2\big]
\end{eqnarray}
where the penalty coefficient is set to $\lambda=10$.

This improved WGAN formulation exhibits high robustness against changing model architecture. The authors tested six different network designs for both $G$ and $D$, which typically break standard GAN training but show stable WGAN training for all cases. Furthermore, WGAN formulation helps achieving better quality samples. Quantitative results are reported by the authors in terms of the \textit{Inception score} \cite{Salimans2016} over CIFAR-10 dataset, which is the most recurrent benchmark for image generation in GAN literature. In the unsupervised setting (without using any class information) they reach the state-of-the-art, while in the supervised case (following the strategy of AC-GAN and without tuning any hyperparameter nor architecture) they reach the second place behind SGAN \cite{Huang2016}\footnote{By simply duplicating the number of feature maps of Gulrajani's networks we found WGAN outperforms SGAN score. See Sec. \ref{sec:results}}. All this advantageous features made --to our knowledge-- WGAN the current standard formulation for adversarial training.

\section{Our Method: Splitting GAN}
\begin{figure*}[h]
 \centering
 \includegraphics[width=13cm,keepaspectratio=true]{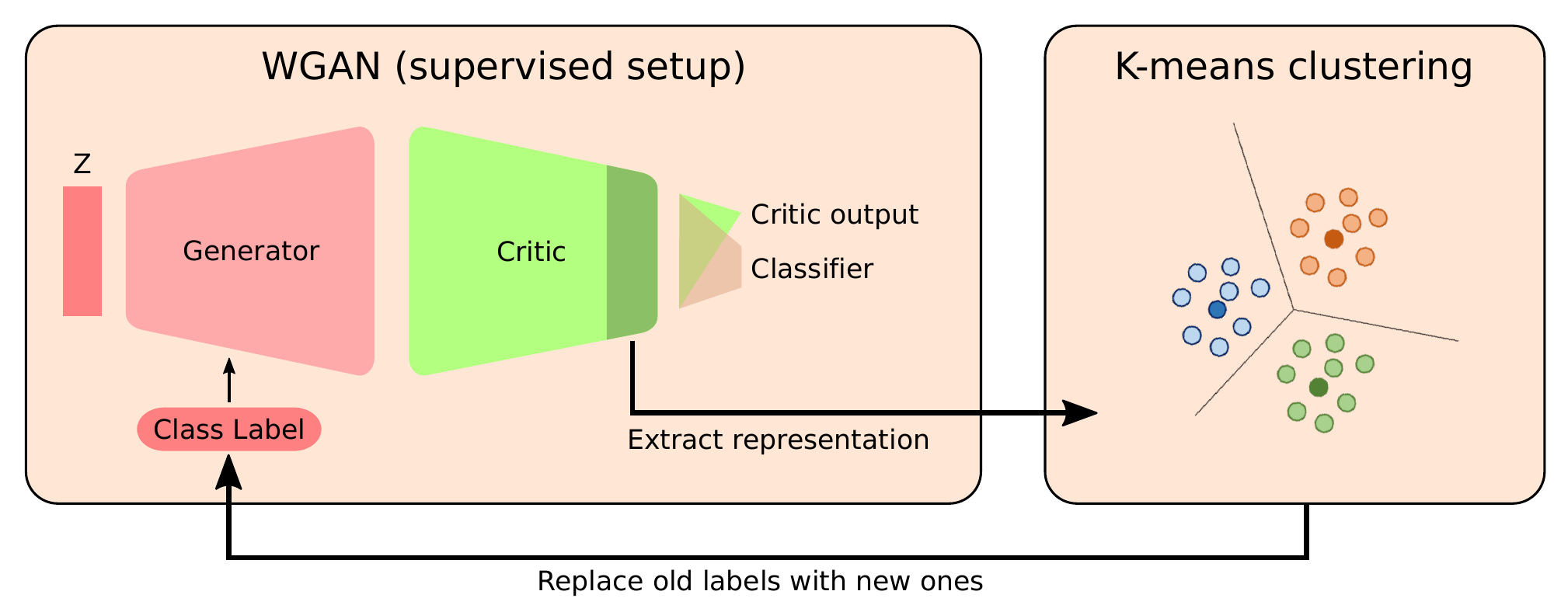}
 \caption{Our Class Splitting GAN proposal is based on generating new classes by clustering in the representation space learned by the critic. This new classes are used in the standard supervised setup of WGAN.}
 \label{fig:model}
\end{figure*}

The main idea is to generate new artificial classes based on the representation learned by the last hidden layer of the critic after enough training iterations. This is done by applying k-means to each class set in this representation space (see Fig. \ref{fig:model}). We divide each set in two clusters only when the class has more samples than a certain threshold. After that, training is resumed replacing the old labels with the new ones for the entire dataset. Algorithm \ref{alg:splitting} resumes the proposed method.

With this procedure we need to make two minor modifications to the model architecture before resuming learning: 
\begin{enumerate}
\item The auxiliary classifier needs to predict a different number of classes, so we extend the last layer of this classifier adding a copy of the weights of the parent class for each child class.
\item In the conditional WGAN-GP implementation \cite{Gulrajani2017github}, the class labels are injected in each batch normalization layer of the generative network by setting a specific gain and bias parameters ($\gamma$ and $b$) for each class. We follow this strategy in our proposal and, for the class splitting, we set the new pair $(\gamma,b)$ for each child class as $\gamma_{child} = \gamma_{father} + \Delta\gamma$ and $b_{child} = b_{father} + \Delta b$, with initialization $\Delta\gamma=0$ and $\Delta b=0$ when the new classes are created. This formulation implies that child classes start both with the father class params and they eventually become different.  
\end{enumerate}

\begin{algorithm}
	\caption{Splitting GAN}
	\label{alg:splitting}
	\begin{algorithmic}
	\REQUIRE A dataset with initial labels (the same label for all the examples in the unsupervised case).
	\REQUIRE $clustering\_iterations$, list of iterations where to make a clustering step.
	\REQUIRE $kmeans\_threshold$, do not divide classes with less samples than this.
    \WHILE{parameters have not converged}
    	\STATE Make a WGAN-GP with auxiliary classifier step, as in \cite{Gulrajani2017}.
    	\IF{current iteration is in $clustering\_iterations$}
      	\FORALL{$class$ in current classes with more than $kmeans\_threshold$ samples}
          	\STATE Propagate through the critic all samples of $class$ up to the last hidden layer.
          	\STATE Normalize these representations.
          	\STATE Apply \emph{K-Means} to the representations in order to obtain two new child classes.
          	\STATE Replace the label $class$ in the dataset with the new child classes.
      	\ENDFOR
     	 
      	\FOR {each new child class, $child$, with parent class $parent$}
          	\STATE In the last layer of the auxiliary classifier, copy its parent parameters.
          	\FOR {each Batch Normalization layer, $i$, of the Generator}
             	\STATE $\gamma_{i,child} = \gamma_{i,parent} + \Delta\gamma_{i,child}$, with initialization $\Delta\gamma_{i,child} = 0$.
             	\STATE $b_{i,child} = b_{i,parent} + \Delta b_{i,child}$, with initialization $\Delta b_{i,child} = 0$.
          	\ENDFOR
      	\ENDFOR
    	\ENDIF
    \ENDWHILE
	\end{algorithmic}
\end{algorithm}

\section{Results}\label{sec:results}
To demonstrate the effectiveness of our proposal, we conducted several experiments with CIFAR-10 \cite{Krizhevsky2009}, a dataset containing 50000 32x32 images corresponding to 10 different classes, and the unlabeled set of STL-10, containing 100000 larger and more diverse images \cite{Coates2011}. We based our model on the improved WGAN algorithm proposed by Gulrajani et al. \cite{Gulrajani2017,Gulrajani2017github}. In all cases, during training, we sample 50000 images from the current model to select the best one so far based on the Inception score. Finally, we sample another 50000 with the best model in order to calculate the reported score, following \cite{Odena2017}.

\subsection{CIFAR-10}

\begin{figure*}
\centering
\includegraphics[width=12cm,bb=0 0 643 320]{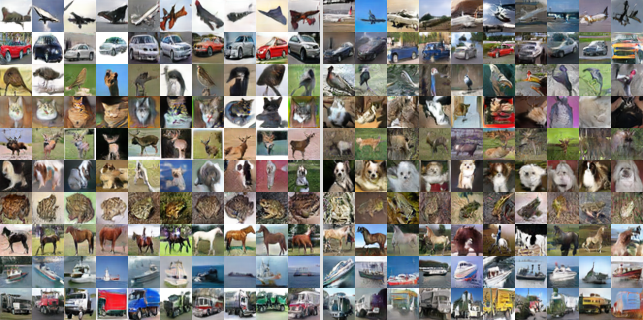}
\caption{Samples generated with our Splitting GAN method with supervised training on CIFAR-10 dataset. Each line has samples of one of the original classes. Each side has samples corresponding to one of the two clusters generated for each class. We use ResNet-B architecture (see text for details).}
\label{fig:cifar10_generated}
\end{figure*}

\begin{table}
\begin{center}
\caption{Unsupervised Inception scores on CIFAR-10}
\begin{tabular}{lc}
\toprule
Method                                    &   Inception Score \\
\midrule
Improved GAN (-L+HA) \cite{Salimans2016} &   $6.86 \pm 0.06$ \\
EGAN-Ent-VI \cite{Dai2017}               &   $7.07 \pm 0.10$ \\
DFM \cite{WardeFarley2017}               &   $7.72 \pm 0.13$ \\
Splitting GAN ResNet-B (ours)              &   $7.80 \pm 0.08$ \\
WGAN-GP ResNet-B                         &   $7.81 \pm 0.10$ \\
WGAN-GP ResNet-A \cite{Gulrajani2017}    &   $7.86 \pm 0.07$ \\
Splitting GAN ResNet-A (ours)              &   $\bm{7.90 \pm 0.09}$ \\
\bottomrule
\end{tabular}
\end{center}
\label{tab:unsup}
\end{table}

\begin{table}
\begin{center}
\caption{Supervised Inception scores on CIFAR-10}
\begin{tabular}{lc}
\toprule
Method                                 &    Inception Score \\
\midrule
Improved GAN \cite{Salimans2016}      &    $8.09 \pm 0.07$ \\
AC-GAN \cite{Odena2017}               &    $8.25 \pm 0.07$ \\
WGAN-GP ResNet-A \cite{Gulrajani2017} &    $8.42 \pm 0.10$ \\
SGAN \cite{Huang2016}                 &    $8.59 \pm 0.12$ \\
WGAN-GP ResNet-B                      &    $8.67 \pm 0.14$ \\
Splitting GAN ResNet-A (ours)           &    $8.73 \pm 0.08$ \\
Splitting GAN ResNet-B (ours)           &    $\bm{8.87 \pm 0.09}$ \\
\bottomrule
\end{tabular}
\end{center}
\label{tab:sup}
\end{table}

\begin{table}
\begin{center}
\caption{Unsupervised Inception scores for STL-10}
\begin{tabular}{lc}
\toprule
Method                                   &   Inception Score \\
\midrule
Original Dataset \cite{WardeFarley2017}  &  $26.08 \pm 0.26$ \\[5pt]
DFM \cite{WardeFarley2017}               &  $ 8.51 \pm 0.13$ \\
WGAN-GP ResNet-A                         &  $ 9.05 \pm 0.12$ \\
Splitting GAN ResNet-A (ours)            &  $\bm{9.50 \pm 0.13}$ \\
\bottomrule
\end{tabular}
\end{center}
\label{tab:sup_stl10}
\end{table}

With CIFAR-10, an unsupervised test was performed starting from all the examples considered as a single class and dividing them into two clusters every approximately 10 epochs.  This was done with the Gulrajani's ResNet architecture without changes (named ResNet-A) and a modified version (ResNet-B) doubling the number of maps in each convolutional layer. A supervised test was also conducted with these two architectures, starting from the original 10 classes of CIFAR and dividing them into two at approximately 20 training epochs. For comparison we also trained the ResNet-B architecture with the original WGAN-GP algorithm. The results are detailed in Table \ref{tab:unsup} and Table \ref{tab:sup}.

\begin{figure}
\centering
\includegraphics[width=6cm,bb=0 0 320 320]{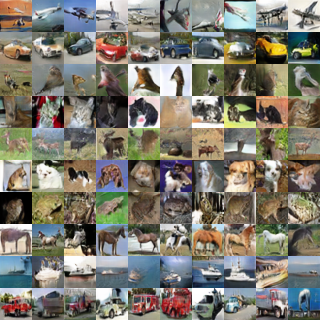}
\caption{Class-conditioned samples generated with WGAN-GP method \cite{Gulrajani2017} over the ResNet-B architecture for CIFAR-10 dataset. Each row has samples of a given class.}
\label{fig:cifar10_wgan_generated}
\end{figure}

\begin{figure*}
 \centering
 \includegraphics[width=12cm,bb=0 0 643 320]{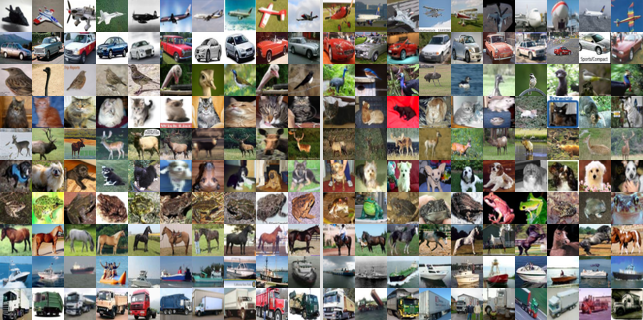}
 \caption{Real CIFAR-10 samples corresponding to the 20 clusters found by our method. Each line is divided in the same way as in Figure \ref{fig:cifar10_generated}.}
 \label{fig:cifar10_clusters}
\end{figure*}

Also for comparison, samples obtained with the proposed method (Figure \ref{fig:cifar10_generated}) and obtained with the WGAN-GP supervised model (Figure \ref{fig:cifar10_wgan_generated}) are shown. Figure \ref{fig:cifar10_clusters} has real samples of CIFAR-10 corresponding to each of the 20 clusters found with our method.

\begin{figure}
 \centering
 \includegraphics[width=6cm, bb=0 0 320 320]{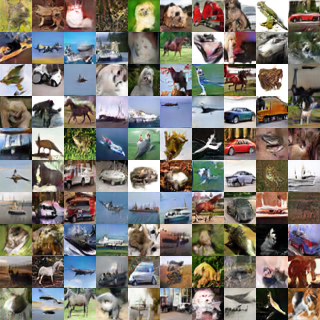}
 \caption{Samples generated with the ResNet-B architecture trained with Splitting GAN over CIFAR-10 without class labels (unsupervised).}
 \label{fig:cifar10_generated_unsupervised}
\end{figure}

\begin{figure}
 \centering
 \includegraphics[width=6cm,bb=0 0 320 800, keepaspectratio=true]{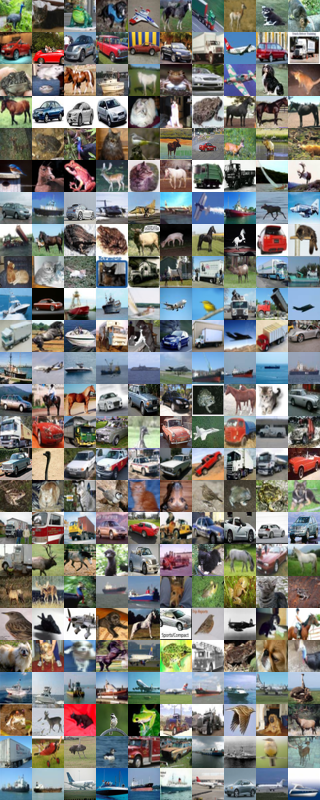}
 \caption{The 25 clusters found in the unsupervised case (real CIFAR-10 samples). Each line has two different clusters.}
\label{fig:cifar10_clusters_unsupervised}
\end{figure}

Figures \ref{fig:cifar10_generated_unsupervised} and \ref{fig:cifar10_clusters_unsupervised} show generated images and clusters found in the unsupervised test.

\subsection{STL-10}
\begin{figure}
 \centering
 \includegraphics[width=6cm,bb=0 0 480 480, keepaspectratio=true]{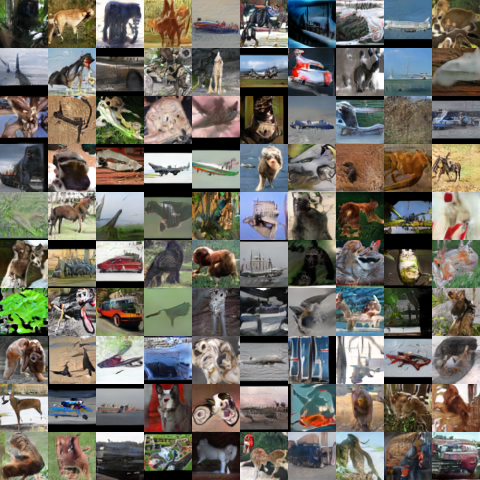}
 \caption{Images generated by the model trained on STL-10.}
\label{fig:stl10_generated}
\end{figure}

\begin{figure}
 \centering
 \includegraphics[width=6cm,bb=0 0 480 720, keepaspectratio=true]{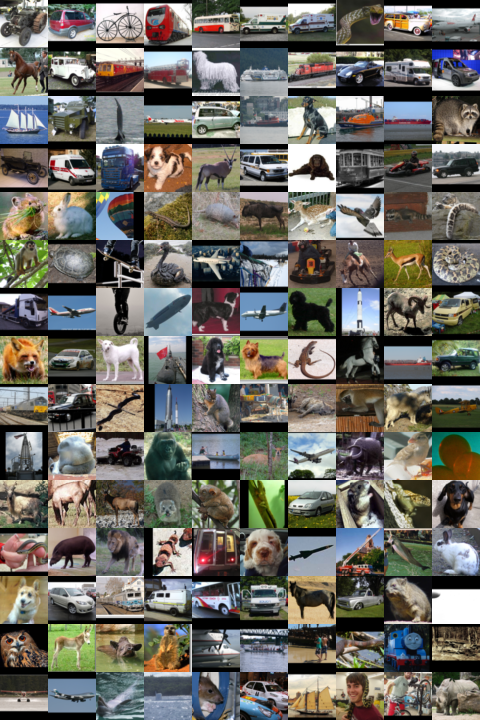}
 \caption{The 15 clusters found by the model (real STL-10 samples).}
\label{fig:stl10_clusters}
\end{figure}

We treat STL-10 in the same way as \cite{WardeFarley2017}. That is, we downsample each dimension by 2, resulting in 48x48 RGB images. We tested our algorithm with the ResNet-A architecture, with the minimum changes necessary for the model to generate 48x48 images. Table \ref{tab:sup_stl10} shows the resulting Inception score.
Figures \ref{fig:stl10_generated} and \ref{fig:stl10_clusters} show the generated images and the clusters found by the method.

\section{Discussion}
Several things can be observed from the results presented in the previous section.
First, regarding the obtained clusterization of the real samples (Figure \ref {fig:cifar10_clusters} for the supervised case and Figure \ref{fig:cifar10_clusters_unsupervised} for the unsupervised one), we can visually find rules that define the vast majority of samples, for at least several clusters.
As an example, in the supervised case (Figure \ref{fig:cifar10_clusters}) we can see in the left side of the fourth row cats looking forward and in the left side of the eighth row horse side views. Compare with cats and horses in several positions corresponding to the clusters in the right side.
In the unsupervised case (Figure \ref{fig:cifar10_clusters_unsupervised}) we can see a tendency to generate clusters for cars, planes or ships, but in general they are much more mixed.

Regarding the generated samples in the supervised case (Figure \ref{fig:cifar10_generated} for our method and Figure \ref{fig:cifar10_wgan_generated} for WGAN-GP), we can see that the class splits allows the model to generate better samples. Not only for the more uniform clusters such as the horse side views or the cats looking forward, but for the whole original class. Compare for example the fourth row (cats) or the eighth row (horses) in Figure \ref{fig:cifar10_generated} with those rows in Figure \ref{fig:cifar10_wgan_generated}, corresponding to the same model trained with WGAN-GP.
Note that these samples do not differ too much from those shown in \cite{Gulrajani2017}. Even in classes where the clustering step does not seem to have found an obvious separation rule, such as cars (second row), a better sample quality can be observed than in the original WGAN-GP.

In the unsupervised case with CIFAR-10 (Figure \ref{fig:cifar10_generated_unsupervised}), although the Inception score is similar than the one obtained by the state-of-the-art so far, the samples generated seem to be of a higher quality. Nevertheless, they do not reach the quality of the generated images in a supervised setting. It is always advisable to start the division into clusters from the predefined classes, if this information is available.

In the case of STL-10 (Figure \ref{fig:stl10_generated}), there is a noticeable difference in the Inception score. The reason of this may be that STL-10 is a much more diverse dataset, so a division into a large number of classes can be beneficial. It should be noted that in this case the state-of-the-art is much further from the actual dataset score than in the case of CIFAR-10.

The success of our Splitting GAN method suggests that reinjecting high level information from critic to the generative model improves sampling quality. This breaks the strictly adversarial training and allows some degree of information sharing between both networks. We believe that this simple (but successful) strategy could inspire a new and better adversarial training formulation where a small amount of high level information directly flows from critic’s last layers to generator input.

\section{Conclusions and Future Work}
In this work we showed that our Splitting GAN method allows generating better images. This can be seen in the results on CIFAR-10 and STL-10 for which clearly better images were obtained. This is supported by an Inception score well above the previous state-of-the-art for both datasets. 

A future direction of research to improve the current Splitting GAN version is oriented to understand how a given model architecture or dataset determines the optimal number of clusters (or classes). Also, clusterization could be enhanced during adversarial training with the addition of an extra loss term like in \cite{Wen2016}.

We are also currently working in a generalization of the Splitting GAN ideas following two paths: First, making the high level information from the critic's representation flows continually to the generator, avoiding special purpose steps at predefined times in the training process. Second, avoiding the hard threshold clustering step and replacing it with some sort of continuous representation capturing the same amount of information.
\section{Acknowledgements}
The authors acknowledge grant support from ANPCyT PICT-2016-4374.

\bibliographystyle{unsrt}
\bibliography{article}

\end{document}